\documentclass[nonacm]{acmart}
\usepackage{dsfont}
\AtBeginDocument{%
  }

\usepackage{algpseudocode}
\usepackage{algorithm}
\begin{document}

\title{Artificial Delegates Resolve Fairness Issues in Perpetual Voting with Partial Turnout}

\author{Apurva Shah}

\affiliation{%
  \institution{Machine Learning Group, Université Libre de Bruxelles }
  \city{Brussels}
  \country{Belgium}}
\affiliation{%
  \institution{AI Lab,Vrije Universiteit Brussel}
  \city{Brussels}
  \country{Belgium}}
\affiliation{%
  \institution{FARI Institute, Université Libre de Bruxelles - Vrije Universiteit Brussel}
  \city{Brussels}
  \country{Belgium}}
\email{apurva.shah@vub.be}

\author{Axel Abels}
\affiliation{%
  \institution{Machine Learning Group, Université Libre de Bruxelles }
  \city{Brussels}
  \country{Belgium}}
\affiliation{%
  \institution{AI Lab, Vrije Universiteit Brussel}
  \city{Brussels}
  \country{Belgium}}
\affiliation{%
  \institution{FARI Institute, Université Libre de Bruxelles - Vrije Universiteit Brussel}
  \city{Brussels}
  \country{Belgium}}
\email{axel.abels@ulb.be}

\author{Ann Nowé}
\affiliation{%
  \institution{AI Lab, Vrije Universiteit Brussel}
  \city{Brussels}
  \country{Belgium}}
\affiliation{%
  \institution{FARI Institute, Université Libre de Bruxelles - Vrije Universiteit Brussel}
  \city{Brussels}
  \country{Belgium}}
  \email{ann.nowe@vub.be}

\author{Tom Lenaerts}
\affiliation{%
  \institution{Machine Learning Group, Université Libre de Bruxelles }
  \city{Brussels}
  \country{Belgium}}
\affiliation{%
  \institution{AI Lab, Vrije Universiteit Brussel}
  \city{Brussels}
  \country{Belgium}}
\affiliation{%
  \institution{FARI Institute, Université Libre de Bruxelles - Vrije Universiteit Brussel}
  \city{Brussels}
  \country{Belgium}}
\affiliation{%
  \institution{Centre for Human-Compatible AI, UC Berkeley}
  \city{Berkeley}
  \country{USA}}
\email{tom.lenaerts@ulb.be}

\begin{abstract}
Perpetual voting addresses fairness in sequential collective decision-making by evaluating representational equity over time. However, existing perpetual voting rules rely on full participation and complete approval information, assumptions that rarely hold in practice, where partial turnout is the norm. In this work, we study the integration of Artificial Delegates, preference-learning agents trained to represent absent voters, into perpetual voting systems. We examine how absenteeism affects fairness and representativeness under various voting methods and evaluate the extent to which Artificial Delegates can compensate for missing participation. Our findings indicate that while absenteeism significantly affects fairness, Artificial Delegates reliably mitigate these effects and enhance robustness across diverse scenarios.
\end{abstract}





\maketitle

\section{Introduction}
How can we design voting systems that remain fair and representative even when not all voters can participate in every round? Perpetual voting considers sequences of decisions made by the same electorate, where fairness must be evaluated over time rather than per decision~\cite{lackner2020perpetual}. A central challenge in this setting is ensuring adequate representation for voters who are repeatedly in the minority. Traditional aggregation rules, such as majority voting or Borda count, fail in this regard: they offer no guarantees of long-term fairness or cumulative influence. In response, methods such as Perpetual Phragm{\'e}n~\cite{lackner2023proportional} and Perpetual Consensus~\cite{lackner2020perpetual} have been proposed to distribute influence more equitably over time. However, they rely on full knowledge of all voters' approval sets, implicitly requiring consistent voter participation, a condition which can be hard to satisfy in real-world contexts.

Real-world elections face various practical constraints---including scheduling conflicts, limited resources, and restricted information access---that inevitably prevent voters from participating consistently.  This partial participation skews the distribution of decision power, as the unrepresented positions of absent voters are not accounted for \cite{bernhagen2007partisan}. Understanding how perpetual voting rules perform under varying degrees of participation is therefore essential for designing robust voting processes.

Existing research on voter absenteeism in large-scale elections often employs statistical techniques, including multiple imputation and regression-based inference, treating missing votes as incomplete data to estimate plausible vote distributions~\cite{bernhagen2007partisan}. Although effective for large populations, these approaches introduce computational complexities and assumptions that diminish their practicality for smaller voting bodies, such as committees or working groups \cite{spratt2010strategies}.

To address absenteeism, liquid democracy~\cite{blum2016liquid, amanatidis2023potential} allows voters either to vote directly or delegate their decision-making power to a trusted representative. Despite its conceptual elegance, liquid democracy depends on voters’ ability to identify suitable delegates, a task often burdensome in practice due to information asymmetries, limited interpersonal trust, or the cognitive complexity of assessing delegate reliability. Furthermore, the transitive delegation mechanism in liquid democracy introduces the risk of power concentration, where a small number of individuals, accumulate disproportionate influence. For instance, if voter A delegates their vote to C, and voter B delegates to A, then C ultimately votes for all three voters. This concentration of voting power can lead to skewed decision-making ~\cite{golz2021fluid, kahng2021liquid}.

In this work, we study a restricted form of liquid democracy wherein absent participants can delegate their votes to personalized Artificial Delegates---agents trained to vote solely on their behalf. Unlike standard liquid democracy, there is no transitive or multi-level delegation: each Artificial Delegate votes only in place of a single voter and holds no independent voting power. These Artificial Delegates are trained using preference learning techniques~\cite{mcfadden1972conditional, train2009discrete} to align with the preferences of the individual they represent, and subsequently vote for them in their absence. 

The ability of these Artificial Delegates to represent voters' preferences depends on how well they can generalize from limited data. Factors such as sparse approval sets, noisy data, and modeling assumptions likely introduce discrepancies between the delegated vote and the voter's true preferences. These imperfections can affect fairness and representativeness of the resulting decisions.  

Understanding such deviations in voting outcomes both for partial turnout and when using Artificial Delegates is critical, and therefore defines the research questions we address in this study:

\begin{enumerate} 
\item How does absenteeism affect the fairness and representativeness of outcomes across multiple state-of-the-art voting methods? 
\item To what extent do artificial delegates mitigate or exacerbate the negative impacts of absenteeism, particularly regarding proportional representation and prolonged voter frustration (dry spells)? \end{enumerate}

We systematically analyze these questions by evaluating outcomes under various voting rules, quantifying fairness through established metrics, such as quota compliance (which assesses whether cohesive groups of voters receive representation proportional to their size), dry spells (the number of consecutive rounds where voters’ preferences remain unrepresented) as well as  Gini influence coefficients (which capture disparities in voter influence)~\cite{lackner2020perpetual}.

We find that partial turnout reduces fairness across perpetual voting rules, resulting in lower quota compliance (voters are less likely to receive proportional representation), longer dry spells (voters go more successive rounds without being satisfied), as well as higher Gini influence coefficients (meaning voter influence is distributed more unequally). In contrast, Artificial Delegates consistently mitigate absenteeism and typically result in fairness metrics which are not significantly different from those obtained with full participation.

In summary, our contributions are as follows:

\begin{itemize} 
\item We analyze the effect of voter absenteeism on perpetual voting outcomes. 
\item We introduce and formalize a novel framework that integrates Artificial Delegates into perpetual voting processes to directly address absenteeism. 
\item We empirically evaluate Artificial Delegates, assessing their influence on various fairness metrics across several state-of-the-art voting rules. 
\end{itemize}

Through this work, we contribute to the development of more robust, equitable, and representative sequential voting systems that better accommodate the reality of inconsistent voter participation.

\section{Perpetual voting, metrics and voting rules}
We consider the setting of perpetual voting introduced by~\cite{lackner2020perpetual}, which addresses decision-making scenarios represented by potentially infinite election sequences. The primary motivation behind perpetual voting is to ensure that the preferences of all voters, including minority groups, are proportionally represented over the long term. By considering historical voter satisfaction, these systems aim to prevent a sustained  "Tyranny of the Majority" --- situations where the majority's preferences consistently outweigh minority groups. Perpetual voting methods typically seek to foster sustained participation from all segments of the electorate by providing a reasonable expectation that their preferences will be reflected in the outcomes at some point in the decision sequence. 

Formally, a perpetual voting instance is a sequence \( I = (E_1, E_2, \dots,  E_T)\) of approval-based elections, where each election \( E_t \) at time step \( t \) consists of a finite set of alternatives \( J_t = \{j_t^1, j_t^2, \dots, j_t^{m_t}\} \), a finite set of voters \( N = \{1, 2, \dots, \mathcal{N}\} \), constant across all elections, and an approval profile \( A_t = (A_t^1, A_t^2, \dots, A_t^{|N|}) \), where \( A_t^i \subseteq J_t \) denotes the subset of alternatives approved by voter \( i \) at time step \( t \). Approval-based elections enable voters to support multiple alternatives simultaneously, offering a simple yet expressive form of preference expression suitable for sequential decisions.

A perpetual voting rule \( \mathcal{R} \) selects a single winning alternative at each step based on the full history of previous elections and outcomes as well as the current election. Formally, for each $t$,

\[
\mathcal{R}: (E_1, j_{1}), (E_2, j_{2}), \dots, (E_{t-1}, j_{{t-1}}), E_t \mapsto j_{t} \in J_t
\]

where \( j_{t} \) is the winning alternative in the current election $E_t$. 

This setting characterizes scenarios where decisions must be made online, with full knowledge of prior elections and their outcomes but without information regarding future alternatives or approval profiles. Consequently, perpetual voting requires balancing of immediate voter satisfaction and strategic long-term fairness considerations~\cite{lackner2020perpetual}.

\subsection{Evaluation Metrics} \label{sec:metrics}
 Consider a perpetual voting history with a voter set \(N\), a set of alternatives \(J_t\) for every timestep $t$, voter approvals \(A_t^n \subseteq J_t\) for every voter $n$ and timestep $t$, and collective decisions \(j_t \in J_t\) at each timestep \(t\). For voter \(n \in N\) and timestep \(t\):

\noindent The \textbf{satisfaction} of voter \( n \in N \) up to timestep \( t \) is the number of collective decisions up to timestep \( t \) approved by voter \( n \):
    \[
    \text{sat}_t(n) = |\{t' \leq t : j_{t'} \in A_t^n\}|
    \]

\noindent  The \textbf{support} of voter \( n \in N \) at timestep \( t \) is:
    \begin{equation}
    \text{supp}_t(n) = \frac{1}{|N|} \max_{j \in A_t^n} |\{m \in N : j \in A_t^m\}|
    \label{eq:support}
    \end{equation}
    This captures the maximum proportion of voters who approve at least one alternative also approved by voter \( n \) at timestep \( t \).
    
The \textbf{quota} of voter \( n \in N \) up to timestep \( t \) is the cumulative support voter \( n \) has received across all timesteps up to \( t \):
    \[
    \text{qu}_t(n) = \sum_{{t'} \leq t} \text{supp}_{t'}(n)
    \]
With these definitions established, we formally define the evaluation metrics as introduced in ~\cite{lackner2020perpetual}:

\paragraph{Longest Dry spell} Longest Dry Spell captures the largest number of consecutive rounds during which an individual voter is unsatisfied with the collective decision. Minimizing this measure ensures voters do not experience excessively long periods without satisfaction. 

A voter experiences a dry spell of length $d$ if the voter agrees with the collective decision at times $t$ and $t+d$, but disagrees with every intermediate decision. Formally, the Longest Dry Spell $D_n$ of voter $n$ is then defined as:

\[
D_n = \max_{t,d}\{\, d \mid   j_{t+k} \notin A_{t+k}^n \, \forall k \in \{1,\dots,d-1\}\}
\]

\paragraph{Gini influence coefficient}
The Gini influence coefficient quantifies inequality in voters' influence over decisions throughout the voting sequence. Influence is defined based on how decisive an individual voter's approval was for each decision: a voter’s influence on a decision is inversely proportional to the number of voters who approved that decision.

Formally, the influence of voter \( n \in N \) across the voting history of length \( T \) is:
\[
\text{infl}_T(n) = \sum_{t=1}^{T}\frac{\mathds{1}\{j_t \in A_t^n\}}{|\{m \in N : j_t \in A_t^m\}|}
\]

Let \(a\) denote the average influence across all voters:
\[
a = \frac{1}{|N|}\sum_{n \in N}\text{infl}_T(n)
\]

The Gini influence coefficient at timestep \(T\) is then defined as:
\[
\text{GiniInf}_T = \frac{1}{2a|N|^2}\sum_{n \in N}\sum_{m \in N}\left|\text{infl}_T(n) - \text{infl}_T(m)\right|
\]

The Gini influence coefficient ranges from 0 (perfect equality in voter influence) to 1 (maximal inequality), offering a detailed measure of fairness in decision-making influence among voters.

\paragraph{Quota Compliance}
Perpetual Lower Quota Compliance (LQC) quantifies how frequently voters meet or exceed their cumulative quota throughout the voting history. Formally, for a voting history of length \(T\), LQC is defined as:
\[
\text{LQC} = \frac{1}{|N|T}\sum_{t=1}^{T}\left|\{n \in N : \text{sat}_t(n) \geq \lfloor \text{qu}_t(n) \rfloor\}\right|
\]

A value of \(\text{LQC} = 1\) indicates perfect compliance, meaning all voters have satisfied their perpetual lower quota at every timestep.

Analogously, Perpetual Upper Quota Compliance (UQC) measures how frequently voters remain within or below their cumulative quota:
\[
\text{UQC} = \frac{1}{|N|T}\sum_{t=1}^{T}\left|\{n \in N : \text{sat}_t(n) \leq \lceil \text{qu}_t(n) \rceil\}\right|
\]

To further clarify deviations from quota compliance, we define complementary measures:

Lower Quota Non-Compliance (LQNC) captures the frequency of voters not meeting their lower quota:
    \[
    \text{LQNC} = 1 - \text{LQC}
    \]

Upper Quota Non-Compliance (UQNC) captures instances of voters exceeding their upper quota:
    \[
    \text{UQNC} = 1 - \text{UQC}
    \]

Finally, we define Quota Compliance (QC) as the proportion of voters simultaneously satisfying both lower and upper quotas:
\[
\text{QC} = 1 - \text{LQNC} - \text{UQNC}
\]

Thus, a higher QC value indicates greater overall adherence to proportional representation, reflecting balanced voter satisfaction relative to their quotas.

\subsection{Perpetual Voting Rules} \label{sec:perpetual}

To address the fairness metrics presented above, several perpetual voting rules have been proposed in the literature~\cite{lackner2020perpetual,lackner2023proportional}. To allow for a comprehensive analysis of absenteeism's impact on perpetual voting, we focus on a subset of these rules, chosen for their representative strengths: Approval Voting as a natural baseline, Perpetual Phragmén and Perpetual Consensus for their strong theoretical guarantees, and Perpetual Quota for its robust empirical performance.

\paragraph{Approval Voting (AV)} Approval Voting~\cite{lackner2020perpetual} provides a straightforward baseline by consistently selecting the alternative approved by the greatest number of voters in each round. Formally, given an approval profile \( A_t \), Approval Voting selects:
\[
j_{t} \in \arg\max_{j \in J_t} |\{ n \in N : j \in A_t^n \}|.
\]

\paragraph{Perpetual Phragm{\'e}n} 

Perpetual Phragm{\'e}n ~\cite{lackner2023proportional} promotes fairness by ensuring that no voter subset consistently dominates decision-making, thereby minimizing dry spells and improving quota compliance (see Section \ref{sec:metrics}). It achieves this by evenly distributing a conceptual "voter load" among voters who approve the selected alternatives.  Formally, each voter \( n \in N \) begins with an initial load \( \ell_n = 0 \). At every round \( t \), the score \( s_j \) of each alternative \( j \in J_t \) is computed as follows:
\[
s_j = \min_{S \subseteq \{n \in N : j \in A_t^n\}} \frac{\sum_{n \in S} \ell_n + 1}{|S|}
\]

The alternative with the minimal score is chosen:
\[
j_t = \arg\min_{j \in J_t} s_j
\]

After selecting alternative \( j_t \), the voter loads are updated. Specifically, the load of each voter in the minimizing set \( S^* \) (which attains the minimum above) is updated:
\[
\ell_n \leftarrow 
\begin{cases}
s_{j_t}, & \text{if } n \in S^*, \\[6pt]
\ell_n, & \text{otherwise.}
\end{cases}
\]

Perpetual Phragm{\'e}n is known to guarantee lower quota compliance, ensuring minority voter groups consistently receive representation proportional to their cumulative support~\cite{lackner2023proportional}.

\paragraph{Perpetual Consensus}  
Perpetual Consensus~\cite{lackner2020perpetual} aims to achieve fairness by dynamically adjusting voter weights based on historical voter satisfaction. Specifically, it redistributes voting influence from voters who have frequently been satisfied toward those who have historically been less satisfied. Initially, each voter \( n \) is assigned equal weight \( \alpha_t^n = 1\). Let $N^+_t(j) = \{ n \in N : j \in A_t^n\; \text{and}\; \alpha_t^n>0 \}$ be the set of voters who approve $j$ and have strictly positive weight at time $t$. At each round \( t \), the alternative selected is the one maximizing the total positive voter weight:
\[
j_t = \arg\max_{j \in J_t} \sum_{n \in N^+_t(j) }\alpha_t^n.
\]

After selecting \( j_t \), voter weights are updated as follows:
\[
\alpha_{t+1}^n \leftarrow 
\begin{cases}
\alpha_t^n + 1 - \frac{|N|}{| N^+_t(j_t) |}, & \text{if } n \in N^+_t(j_t) , \\[6pt]
\alpha_t^n + 1, & \text{otherwise}.
\end{cases}
\]

While Perpetual Consensus does not explicitly minimize dry spells, it effectively promotes overall proportional representation by complying with perpetual upper quota conditions, thus preventing any voter group from excessively dominating decision-making beyond their fair cumulative share~\cite{lackner2023proportional}.

\paragraph{Perpetual Quota}
Perpetual Quota~\cite{lackner2020perpetual} directly aims to minimize deviations from proportionality by selecting alternatives that align voter satisfaction closely with cumulative voter quotas. It specifically prioritizes voters whose historical satisfaction is below their cumulative quota.

Formally, at each round \( t+1 \), Perpetual Quota selects the alternative \( j_{t+1} \) maximizing the total positive deviation from voters' cumulative satisfaction relative to their quota:
\[
j_{t+1} = \arg\max_{j \in J_{t+1}} \sum_{n \in N : j \in A_{t+1}^n} 
\max\left(0,\;\text{qu}_{t+1}(n) - \text{sat}_t(n)\right).
\]

Thus, Perpetual Quota explicitly reduces proportionality differences and improves quota compliance, leading to lower inequalities in voter satisfaction over time.

\section{The Problem of Missing Voters} \label{sec:imputation}

In practice, voters may not participate consistently in every round due to factors outside of their control such as illness, emergencies, or systemic barriers. Increasing attention has therefore been devoted to developing methods for imputing preferences of missing voters, particularly those who abstain or submit incomplete ballots\cite{spratt2010strategies,doucette2014imputation, bernhagen2010missing,ren2020missing,liu2014using}. These imputation techniques aim to estimate how non-voters or partially participating voters would have voted, thereby treating voter absenteeism as a missing data issue within electoral datasets\cite{spratt2010strategies,doucette2015resolving, liu2014using,bernhagen2010missing}. 

One commonly employed approach is multiple imputation (MI), which fills in missing preferences by generating plausible values based on probabilistic distributions. MI has proven useful in analyzing the potential partisan impacts of low voter turnout in electoral outcomes~\cite{bernhagen2007partisan,bernhagen2010missing, liu2014using}.  Besides traditional statistical techniques, researchers have increasingly explored machine learning algorithms for predicting absent voters' preferences~\cite{spratt2010strategies,doucette2014imputation,doucette2015resolving}.  These methods often rely on identifying patterns in the preferences of voters who did participate and using these patterns to infer the likely preferences of non-voters based on their available characteristics or any partial preferences they might have provided. Other approaches include methods like Coarsened Exact Matching (CEM), which imputes the voting choices of abstainers based on the voting choices of voters who are exactly similar to them on a set of relevant covariates~\cite{bol2021s}. 

By attempting to account for the estimated preferences of absent voters, these methods more accurately reflect the true distribution of preferences within the entire eligible voting population~\cite{spratt2010strategies,doucette2014imputation, doucette2015resolving, bernhagen2010missing}. For instance, research has shown that MI can be a valuable tool for estimating the impact of turnout on election results~\cite{bernhagen2007partisan,bernhagen2010missing}. Similarly, machine learning-based approaches have demonstrated the ability to recover the correct winner in elections with partial voter preferences with a high degree of probability~\cite{doucette2014imputation}. 

However, these imputation techniques rely on certain assumptions that may not hold true in real-world scenarios~\cite{ren2020missing}. For instance, demographic similarity may not reliably predict voting preferences, and machine learning methods require substantial data to train predictive models effectively~\cite{doucette2014imputation}.

Beyond simple votes, \citet{lang2012winner} analyzes how to aggregate incomplete orders over sets of candidates in sequential majority voting, where outcomes are determined by sequences of pairwise majority votes. It studies how to identify winners when agents provide only partial rankings, examining the computational difficulty of doing so and introducing efficient heuristics to approximate the set of winners. We in contrast focus on perpetual voting and its challenges in terms of fairness.

To reduce the cognitive load on voters, \citet{halpern2023representation} propose querying voters for a partial approval profile over the candidates and applying voting rules. To support their approach,  they conduct an empirical study using open data from Polis and data generated from Reddit discussions. In follow-up work \cite{halpern2024computing}, they also extend this approach to rank-based settings.  However, these approaches focus on single-election settings with partial candidate votes (or rankings). In contrast, we investigate perpetual voting scenarios, where some voters’ preferences may be entirely absent in certain rounds. This requires inferring long-term voter preferences, rather than overcoming partial, one-time votes (or rankings) explored in these previous works.

In the perpetual voting setting, \citet{chandak2024proportional} apply preference learning to estimate country-specific preferences using the Moral Machine dataset \cite{awad2018moral}. They construct representative models for each country, trained in advance on extensive survey data, and use these models to simulate perpetual voting scenarios. The outcomes of those perpetual votes are then shown to be more representative than those that would have been taken by a single global model.  Crucially, these artificial delegates vote in isolation, not alongside real voters. In contrast, we use preference learning to dynamically represent absent voters within ongoing perpetual voting, focusing on how outcomes differ from those that would have occurred under full participation.

Neither their work nor the broader perpetual voting literature (\cite{lackner2020perpetual,lackner2023proportional}) explicitly address voter dropout, leaving an interesting gap in our understanding of how inconsistent participation impacts fairness over time. This leaves an open question regarding the long-term fairness implications of inconsistent participation. Our work directly targets this gap by (i) evaluating how absenteeism affects outcomes in perpetual voting, and (ii) introducing methods that exploit historical voting behavior in sequential collective decision-making to automatically represent absent voters through Artificial Delegates.

\section{Methods} \label{sec:methods}
In addition to investigating the impact of absenteeism on perpetual voting, we also study whether incorporating learned preferences can mitigate these effects, thereby improving fairness. In contrast to the MI works presented in \autoref{sec:imputation}, our aim here is to substitute absent voters during the voting process, rather than analyzing the impact of their absence after voting is completed.

A naive perspective holds that voters who are absent have voluntarily forfeited their voting rights, and thus, their absence can be interpreted as tacit consent to any outcome reached without their input. However, this view risks unfairly under-representing individuals whose absence stems from uncontrollable circumstances, such as illness, emergencies, or systemic barriers.

A seemingly fairer baseline treats absent voters’ approval sets as empty:

\begin{equation} \hat{A}_t^n = \begin{cases} A_t^n & \text{if } n \text{ is present for election } t \\ \emptyset & \text{otherwise} \end{cases} \label{eq:approval_missing}
\end{equation}

This formulation results in absent voters receiving zero support in any election they miss. Since any participating voter would at least approve of their own preferences, full participation guarantees support of at least $1/|N|$ per voter. Consequently, absent voters’ supports---and by extension, quota-based fairness guarantees---are systematically underestimated. Although satisfaction is also underestimated---since empty approval sets imply dissatisfaction---the symmetry is misleading. In practice, an absent voter might have approved the outcome and thus gained  both support and satisfaction, neither of which are recorded.

These distortions propagate through voting rules that rely on support and satisfaction, such as Perpetual Quota (see Section \ref{sec:perpetual}). In particular, absent voters are excluded from both support and quota calculations. As a result, not only are absent voters misrepresented, but present voters also receive less support than they would have if absentees were not treated as having empty approval sets. However, their satisfaction is not underestimated in the same way, causing Perpetual Quota to overestimate voter satisfaction relative to their quota. This mismatch is likely to consistently lead to under-satisfaction. To mitigate this mismatch, we substitute the denominator in \autoref{eq:support} by the number of present voters. 

More sophisticated rules like Perpetual Phragmén and Perpetual Consensus are also likely to be affected.

Consider first Perpetual Consensus (see \autoref{sec:perpetual}), a rule that operates on a virtual consensus model by decreasing voter weight by 1 when all voters are satisfied. In practice, satisfied voters receive a penalty subtracted from a base increment. When absenteeism leads to empty approval sets, absent voters are always marked as unsatisfied. As a result, their weights increase by one unit each round without incurring penalty, leading to inflated weights relative to active voters. This distorts the consensus dynamics, granting disproportionate influence to consistently absent participants.

A parallel issue arises in Perpetual Phragmén (see \autoref{sec:perpetual}), where voter loads represent cumulative support. Typically, voters share the load cost of an elected alternative. However, absent voters with empty approval sets are exempt from sharing that load. Over time, this results in lower accumulated loads. Upon returning, such voters hold disproportionately low load levels and thus wield excessive influence relative to consistently active voters, undermining the rule’s core load-balancing mechanism.

To mitigate these systematic distortions of voting dynamics while preserving proportionality and voter intent, we introduce a novel approach inspired by Liquid Democracy~\cite{blum2016liquid}: representing absent voters through \emph{Artificial Delegates} whose preferences approximate those of the absent individuals. These Artificial Delegates are constructed using preference learning methods and vote on behalf of absent voters, based on previously observed individual preference patterns.

Unlike the traditional imputation approaches that identify similar voters to estimate absentee preferences (see Section \ref{sec:imputation}), we propose to take advantage of the sequential nature of perpetual voting in order to identify individual preferences more effectively.

\subsection{Learning to Represent Absent Voters}
We consider voting sequences wherein, at each timestep \(t\), voters observe a set of alternatives, each characterized by context features (e.g., cost-effectiveness, feasibility, or impact).  Formally, alternatives at timestep \(t\) are characterized by context vectors \((\mathbf{x}_t^j)_{j=1}^{|J_t|}\), where each \(\mathbf{x}_t^j \in \mathbb{R}^d\) encodes relevant features or objectives. For instance, each alternative may represent trade-offs among conflicting objectives like monetary cost versus efficiency.

\subsubsection{Random Utility Model} \label{sec:RUM}
We assume participants make decisions according to a random utility model. Given the feature vector $\vec{x}_t^j$ describing option $j$ at time $t$, each individual assigns weights to objectives (e.g., cost-effectiveness, feasibility) based on a personal preference vector $\mathbf{W}^n \in \Delta(d)$, where $\Delta(d)$ denotes the $d$-dimensional standard simplex. The utility of an option is defined as $\langle \vec{x}_t^j, \mathbf{W}^n \rangle$. An option is included in the voter's approval set if its utility exceeds a voter-specific threshold $\tau_n$.

Although voters aim to maximize personal utility, real-world decision-making may be noisy due to cognitive limitations or uncertainties. As in~\cite{ben1995discrete,cascetta2001random}, judgment errors, capturing either uncertainty about the problem's characterization $\vec{x}_t^j$ or variability in preferences $\mathbf{W}^n$ can be equivalently modeled by varying the strength of some random  noise $\varepsilon^n_t$: 

$$ j \in A^n_t \iff \langle \vec{x}_t^j, \mathbf{W}^n \rangle + \varepsilon^n_t > \tau_n$$

This formulation captures the probabilistic nature of human choice, including potential inconsistencies and judgment errors.

\subsubsection{Preference Learning} \label{sec:preference}

To estimate $\mathbf{W}^n$ for an absent voter based on a history of approval sets $\{A_t^n\}_t$, we formulate a maximum likelihood estimation problem under the logistic model. Letting $\vec{x}_t^j$ represent the features of option $j$ at time $t$ and $y_{t,j}^n = \mathds{1}\{j \in A_t^n\}$, we maximize the log-likelihood of observed approvals subject to the simplex constraint on $\mathbf{W}^n$:
\[
\begin{aligned}
\max_{\mathbf{\hat{W}}^n \in \Delta(d),\ \hat{\tau}_n \in \mathbb{R}} \quad & \sum_{t,j} \left[ y_{t,j}^n \cdot \log \sigma\left( \langle \vec{x}_t^j, \mathbf{\hat{W}}^n \rangle - \hat{\tau}_n \right) + (1 - y_{t,j}^n) \cdot \log \left(1 - \sigma\left( \langle \vec{x}_t^j, \mathbf{\hat{W}}^n \rangle - \hat{\tau}_n \right) \right) \right]
\end{aligned}
\]
where $\sigma(z) = 1 / (1 + e^{-z})$ denotes the logistic function.

Note that this is equivalent to performing logistic regression with a simplex constraint on the weight vector.

This constrained optimization problem can be solved using convex programming techniques, ensuring that the learned preferences $\hat{\mathbf{W}}^n$ remain within the simplex and are interpretable as weights over objectives. This formulation enables preference recovery for absent voters based solely on their historical approval behavior, without requiring explicit pairwise comparisons or rankings.

The inferred preferences enable Artificial Delegates to represent absent voters by predicting their approval sets: 

\begin{equation}
\hat{A}_t^n = \left\{ j \in [m] \;\middle|\; \langle \vec{x}_t^j, \mathbf{\hat{W}}^n \rangle \geq \hat{\tau}_n \right\} \label{eq:approval_learned}
\end{equation}

At each timestep $t$, voting rules incorporate these predicted approval sets as substitutes for missing voters.

\subsection{Outcome Alignment} \label{sec:overlap}

The primary aim of Artificial Delegates is to yield outcomes that are both fairer than those produced under partial turnout and not drastically different from outcomes that would emerge without delegation. Ideally, the delegated outcome should remain consistent with the preferences of either (i) the participating voters or (ii) the full electorate, had everyone voted.

To quantify this alignment, define:

 \( j_t^{\text{delegate}} \), the election outcome at time \( t \) using estimated approval sets (see Section \ref{sec:preference});
 \( j_t^{\text{full}} \), the outcome using the full approval sets;
\( j_t^{\text{partial}} \), the outcome using only the observed approval sets, with non-participating voters assumed to approve no candidates (see \autoref{eq:approval_missing}).

We define overlap over $k$ rounds as the proportion of rounds where the delegate outcome matches either the full turnout or the partial turnout result:

\[
\text{overlap} = \frac{1}{k} \sum_{t=1}^{k} \mathds{1}\{ j_t^{\text{delegate}} = j_t^{\text{full}} \; \text{or} \; j_t^{\text{delegate}} = j_t^{\text{partial}}  \}
\]

This metric reflects how often delegation preserves agreement with either the full electorate or the actual voters.

\section{Experimental Design}\label{sec:experiments}
To evaluate how absenteeism affects perpetual voting, we simulate\footnote{To ensure reproducibility, the code to reproduce these results will be made publicly available upon publication.} perpetual voting scenarios in line with previous works~\cite{lackner2020perpetual,lackner2023proportional}. Unless stated otherwise, all results are averaged over 100 independent simulations, each consisting of 50 voting rounds. Each round, $20$ voters choose among 5 alternatives, each characterized by a 5-dimensional feature vector $\vec{x}_t^j \sim \mathcal{U}([0,1]^5)$, independently sampled for each round.


\begin{figure}[h]
\centering
\includegraphics[width=1\textwidth]{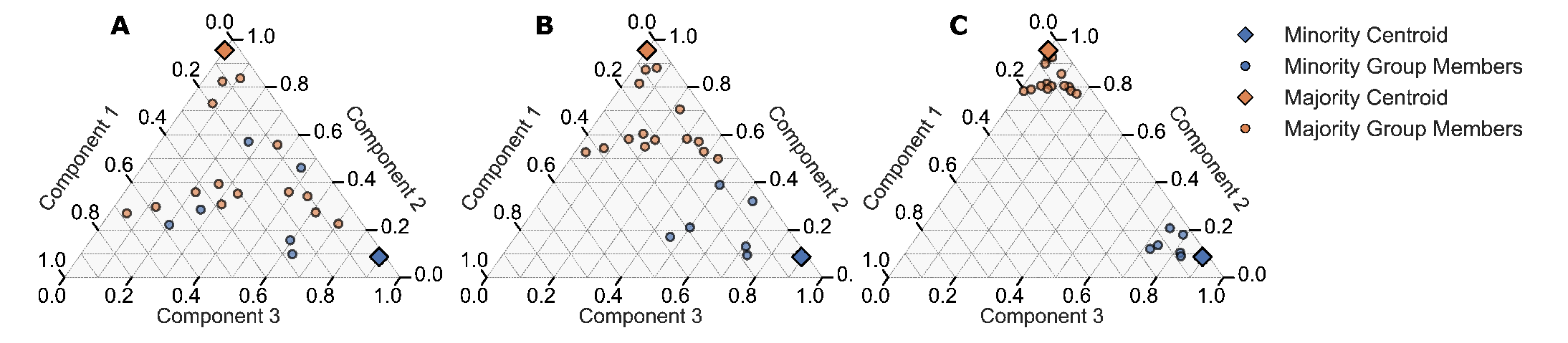}
\caption{Illustration of the impact of cluster densities on voter spread ($p$, see \autoref{sec:experiments}) in the $3$-dimensional simplex. For each simulation, we sample a majority and minority centroid (diamonds), and then sample individual voter weights (circles) around those centroids.  Increasing the cluster density from $0.2$ (\textbf{A}), through $0.5$ (\textbf{B}), and to $0.8$ (\textbf{C}) concentrates voters around the centroids, emulating polarization. Although this figure uses a 3-dimensional simplex for visualization, the experiments described in Section \autoref{sec:experiments} are performed in a 5-dimensional simplex.
}
\Description[Three ternary plots showing voter spread at different cluster densities.]{Three ternary plots show the distribution of majority and minority voter groups at cluster densities 0.2, 0.5, and 0.8. Each plot displays centroids and individual voters. As density increases, voters cluster more tightly around centroids, illustrating growing polarization.}
\label{fig:densities}
\end{figure}

We assume voters follow the random utility model described in  Section \ref{sec:RUM}, and, as illustrated in \autoref{fig:densities}, sample their preferences as follows: first, for each group we sample a preference centroid
$\mathbf{W}^{\text{centroid}}$ from the $5$-dimensional standard simplex following a Dirichlet distribution $\text{Dir}(\{0.2,...,0.2\})$. Then, each voter's preference vector is drawn by perturbing their group's centroid: 
$ p\mathbf{W}^{\text{centroid}} + (1-p)\mathbf{W}^{\text{offset}}$, where $p$ (by default $0.5$) controls the cluster density and $\mathbf{W}^{\text{offset}}$ is sampled from the same Dirichlet distribution. 
The resulting voter preferences form two distinguishable but possibly overlapping populations on the simplex. As \autoref{fig:densities} illustrates, increasing the cluster density leads to increased polarization of the voters. 

As in~\cite{lackner2020perpetual,lackner2023proportional}, voters are divided into a majority group of size $14$ and a minority group of size $6$. When evaluating different group sizes we maintain this ratio, e.g., $7$ to $3$ for voter sets of size $10$.

In addition to preferences, approval sets are controlled by $\tau=0.05$, the threshold for inclusion, as well as $\beta=0.01$, the scale of the noise. In the case of $5$ alternatives, these values result in approval sets of size $\approx 2$, similar to the experimental setups of \citet{lackner2020perpetual} and \citet{lackner2023proportional}. 

In the full turnout scenario, these approval sets are then aggregated into a collective decision using the perpetual voting rules discussed in Section \ref{sec:perpetual}. 

To understand the effect of absenteeism, and the potential benefits of Artificial Delegates, we evaluate two more participation modes: (i) partial turnout, in which case absent voters are assumed to have empty approval sets (see \autoref{eq:approval_missing}), and (ii) delegated participation, wherein absent voters are represented by learning delegates, which involves the preference learning detailed in Section \ref{sec:preference}. 

\section{Results and Discussion}
\begin{figure}[h]
\centering
\includegraphics[width=1\textwidth]{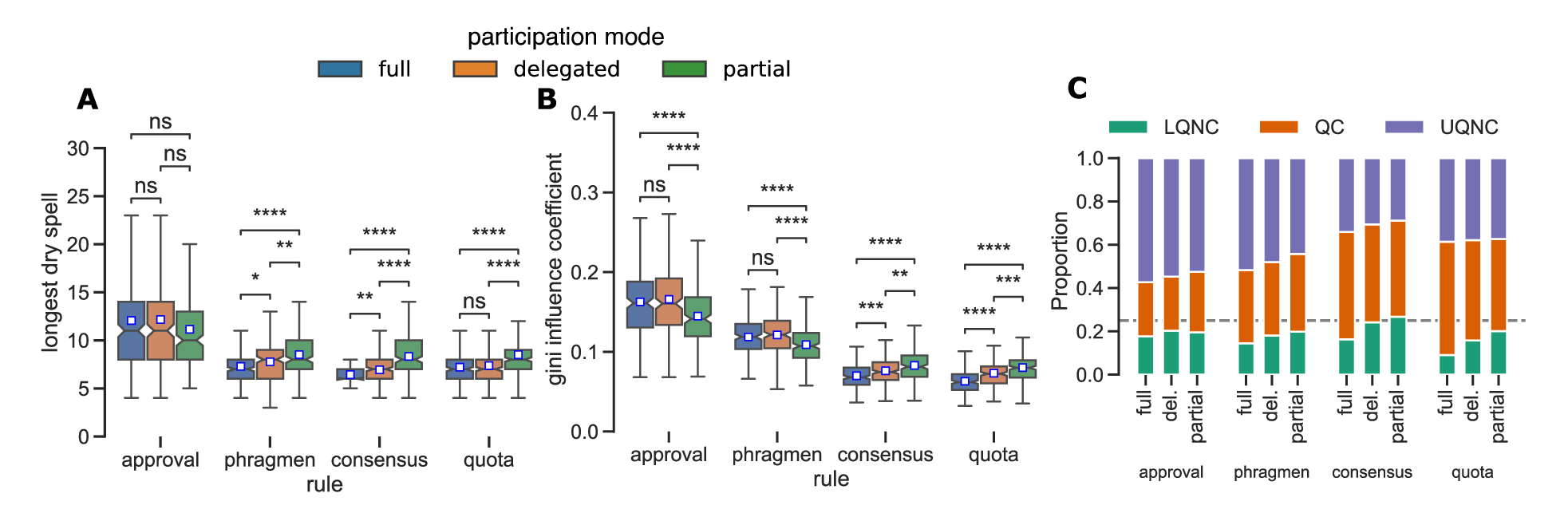}
\caption{Comparison of voting rules across three participation modes --- \textit{full}, \textit{partial}, and \textit{delegated} --- for different evaluation metrics. Parameters are set to their defaults as defined in \autoref{sec:experiments}. \textbf{A,B} Each boxplot represents the distribution for a specific rule over a given voter set type. Notches indicate confidence intervals around the median, and squares denote the mean. Statistical differences across participation modes for each rule are tested using Kruskal–Wallis tests, followed by post-hoc Dunn tests. Significance levels:
ns ($p > 0.05$), * ($p \leq 0.05$), ** ($p \leq 0.01$), *** ($p \leq 0.001$), **** ($p \leq 0.0001$). \textbf{C} Distribution of LQNC, QC, and UQNC per rule and participation modes.
}
\Description[Two Boxplots and a stacked bar chart comparing fairness metrics across voting rules and participation modes.]{Three subplots, from left to right, compare Longest dry spell, Gini Influence coefficient and Quota Compliance respectively across four voting rules (Approval Voting, Perpetual Phragmén, Perpetual Consensus, and Perpetual Quota) under three voter participation types: full, delegated, and partial.
Subplot A compares the Longest Dry Spell across rules and participation modes. Under the Approval Voting rule, no significant differences are observed between participation modes. In contrast, for Perpetual Phragmén, Perpetual Consensus, and Perpetual Quota, statistically significant differences emerge: full participation results in the least longest dry spell, followed by delegated, with partial participation leading to the highest longest dry spell. An exception occurs under Perpetual Quota, where the difference between full and delegated turnout is not statistically significant.
Subplot B depicts the Gini Influence Coefficient under the same conditions. For Perpetual Consensus and Perpetual Quota, there are significant pairwise differences across participation types, with full turnout yielding the lowest Gini influence coefficient, followed by delegated turnout and then partial turnout. However, under Approval Voting and Perpetual Phragmén, no significant difference is observed between full and delegated turnout, and both exhibit higher Gini coefficients than partial turnout.
Subplot C is a stacked bar graph illustrating distribution of LQNC, QC and UQNC across Approval Voting, Perpetual Phragmén, Perpetual Consensus, and Perpetual Quota. For Approval voting and Perpetual Phragmén, UQNC substantially dominates, followed by QC, with LQNC having the smallest proportion; distributions show little variation across participation modes. In contrast, for Perpetual Consensus and Perpetual Quota, QC moderately exceeds UQNC, with LQNC remaining the smallest. These rules also show a slight increase in QC—and consequently a decrease in LQNC—under full participation compared to other participation modes. 
}
\label{fig:boxplots_composition}
\end{figure}

\subsection{Impact of voter turnout on fairness in perpetual voting}\label{sec:impact}
As a first step towards understanding the impact of absenteeism, and the potential benefits of learning Artificial Delegates, we first present summarized results in \autoref{fig:boxplots_composition}. These plots compare the performance of various voting rules for different voter participation modes (full, partial, or delegated), and for different metrics (see Section \ref{sec:metrics}). 

Although the metrics capture distinct fairness dimensions, a consistent trend emerges: partial turnout generally leads to lower fairness, reflected in higher values for these metrics. Among the voting rules, Perpetual Phragmén is least adversely affected by absenteeism (excluding approval voting). While it exhibits longer dry spells---periods where a voter receives no beneficial outcomes--- absenteeism reduces its Gini influence coefficient, signaling greater equality in voter influence. This should however be nuanced by the observation that its Gini influence coefficient is typically much higher than Perpetual Consensus' or Perpetual Quota's. These two methods are significantly harmed by partial turnout for all metrics. Note that, while its design resembles Perpetual Consensus, Perpetual Phragmén differs in its distribution of the cost of approval. In particular, Perpetual Consensus divides the cost of approval equally over approving voters, while Perpetual Phragmén assigns more cost to low-load voters. 

Surprisingly, the majority vote can benefit from partial turnout. We believe this is because partial turnout can skew the vote towards the minority in the instances where many majority members are absent simultaneously. Hence, while full turnout would always result in a majority win, absenteeism sometimes allows for a minority win.

Bars for approval voting in \autoref{fig:boxplots_composition}.C display, as expected, a proportion of over-representation (UNQC) and under-representation (LQNC) in line with the proportions of the minority and majority groups. 
Note that, in line with \citet{lackner2023proportional}'s analysis, Perpetual Phragm{\'e}n ensures under-representation is limited, but provides no guarantees in terms of over-representation. We observe similar empirical results here, as Perpetual Phragm{\'e}n displays little LQNC, meaning voter's satisfaction is typically lower bounded by their quota: they are at least as satisfied as they should be. In contrast, it displays high UQNC, meaning voters often get more representation than their quota suggests they should get. In general, partial turnout has little impact on QC. We hypothesize this is because absenteeism is as likely for both minority and majority groups. As a result, both groups tend to miss a similar portion of voters, leading to neither group being disproportionally present or absent. 

\paragraph{Artificial Delegates} Having established that absenteeism typically harms perpetual voting rules, we now evaluate to what degree Artificial Delegates effectively represent their delegators. Towards this, we observe that \autoref{fig:boxplots_composition}.A and \autoref{fig:boxplots_composition}.B show that, while averages do slightly increase, representation through Artificial Delegates results in longest dry spells and Gini influence coefficients that are typically significantly better than those obtained with full turnout. This suggests that the artificial delegates are effective at representing missing voters. 

In terms of QCs, \autoref{fig:boxplots_composition}.C shows that delegates result in similar QC for approval voting and Perpetual Phragmén, but slightly reduce QC for Perpetual Consensus and Perpetual Quota when compared to full turnout.

Note that these results only capture one specific configuration, and the effectiveness of Artificial Delegates could be impacted by several factors, which we now explore.

\subsection{How are results affected by variations in the main parameters}

While our previous discussion summarizes the effects of absenteeism, we now explore the effects various parameters have on Gini influence coefficients and Longest Dry Spells.

\paragraph{Number of Options}

\begin{figure}[ht]
\centering
\includegraphics[width=1\textwidth]{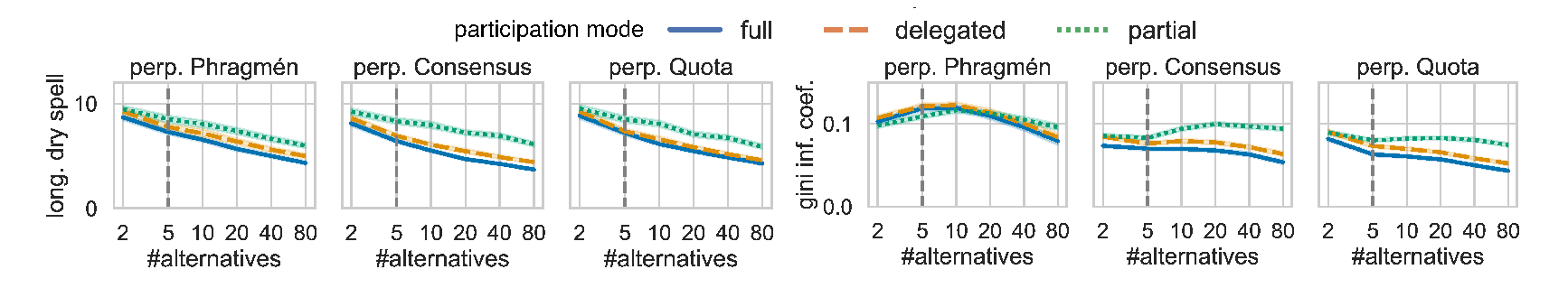}
\caption{Effects of the number of alternatives on Gini influence coefficients and longest dry spells for three perpetual voting rules. Vertical dashed lines mark the default value, as used for \autoref{fig:boxplots_composition}. All other parameters are set to their defaults as defined in \autoref{sec:experiments}. 
}
\Description[Line plots show how increasing alternatives affects fairness metrics across voting rules and participant modes.]{ The first three plots show that, for all three voting rules, an increase in the number of alternatives reduces the Longest Dry Spell under full, delegated, and partial turnout conditions. Partial turnout consistently yields the highest longest dry spell, followed by delegated turnout, with full turnout resulting in the least. The subsequent three plots illustrate corresponding patterns in the Gini Influence Coefficient. A general decline is observed across all rules as the number of alternatives increases. Partial turnout typically exhibits the highest Gini Influence, followed by delegated and full turnout. An exception is noted in Perpetual Phragmén, where partial turnout initially yields the lowest Gini Influence coefficient up to approximately 20 alternatives, after which it surpasses the other modes}

\label{fig:variants_alternatives}
\end{figure}
The number of available alternatives significantly affects fairness, as shown in \autoref{fig:variants_alternatives}. Fairness improves across all methods as the number of alternatives increases. We hypothesize this is due to larger approval sets, which lead to greater overlap among voters’ approval sets, making it easier to identify broadly satisfactory options. Notably, our Artificial Delegates system closely tracks full participation, while the gap in terms of Gini influence coefficients widens for partial turnout.

\paragraph{Number of Voters}

\begin{figure}[h]
\centering
\includegraphics[width=1\textwidth]{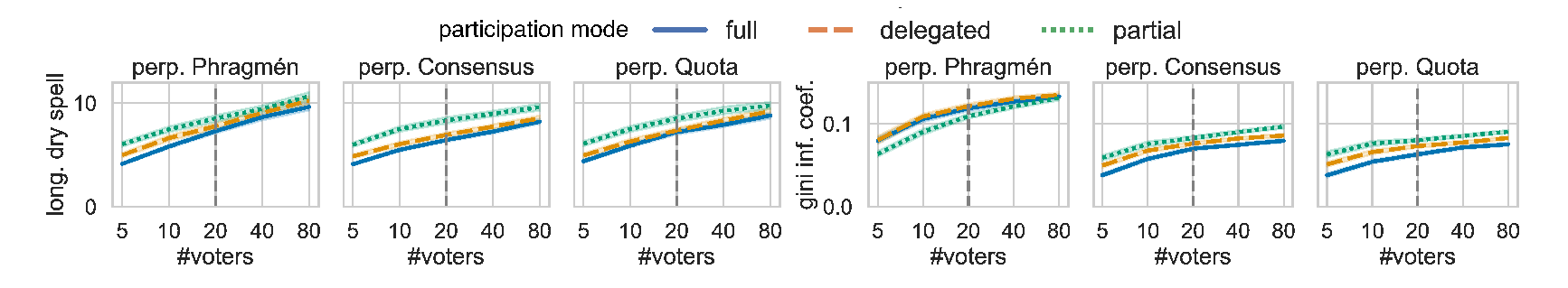}
\caption{Effects of the number of voters on Gini influence coefficients and longest dry spells for three perpetual voting rules. Vertical dashed lines mark the default value, as used for \autoref{fig:boxplots_composition}. All other parameters are set to their defaults as defined in \autoref{sec:experiments}. 
}
\Description[Six Line plots show how fairness metrics change with the number of voters across voting rules, and participation modes.]{Line plots show that as the number of voters increases, both the Longest Dry Spell and Gini Influence Coefficient generally increase. Partial turnout typically exhibits the highest values for these metrics, followed by delegated and full turnout. An exception occurs in Perpetual Phragmén, where partial turnout has the lowest Gini Influence Coefficient among the three participation modes.}
\label{fig:variants_voters}
\end{figure}

As shown in \autoref{fig:variants_voters}, increasing the number of voters generally leads to reduced fairness, as reflected by higher Gini influence coefficient and Longest Dry Spell values. This results from the greater diversity of preferences in larger electorates, which cannot all be accommodated simultaneously. As a consequence, voters wait longer to be represented (longer dry spells) and experience more unequal satisfaction (higher Gini influence coefficient).

Interestingly, for Artificial Delegates, the gap to full participation narrows as the number of voters increases. This likely occurs because absenteeism becomes more uniformly distributed in a larger population, reducing its biasing effect and aligning the outcome more closely with that of full participation.

\paragraph{Impact of Absenteeism}

\begin{figure}[ht]
\centering
\includegraphics[width=1\textwidth]{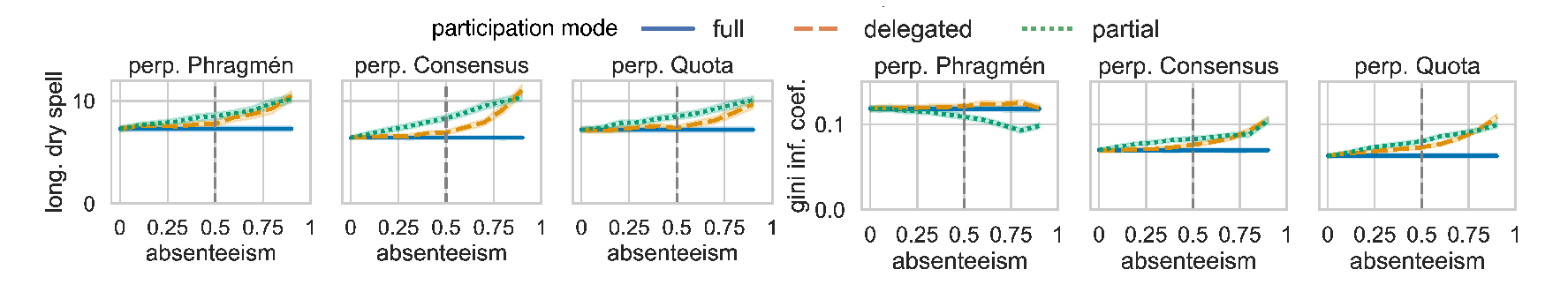}
\caption{Effects of absenteeism on Gini influence coefficients and longest dry spells for three perpetual voting rules. Vertical dashed lines mark the default value, as used for \autoref{fig:boxplots_composition}. All other parameters are set to their defaults as defined in \autoref{sec:experiments}. 
}
\Description[Six line plots show how change in absenteeism affects fairness metrics across voting rules and participant modes]{ Line plots show that as absenteeism increases, values of both fairness metrics, Longest dry spell and Gini influence coefficient, tend to increase for delegated and partial turnout, while remaining stable under full turnout. Partial turnout generally has the highest values, followed by delegated, with full turnout having the lowest. An exception occurs in Perpetual Phragmén, where partial turnout has the lowest Gini Influence Coefficient among the three participation modes.}
\label{fig:variants_absenteeism}
\end{figure}
We evaluate the effects of absenteeism by systematically varying the probability that a voter abstains, as shown in \autoref{fig:variants_absenteeism}. As absenteeism increases, fairness metrics deteriorate more rapidly under partial participation. This is especially pronounced for Perpetual Quota and Perpetual Consensus, where both the longest dry spells and Gini influence coefficients rise sharply. In contrast, delegated participation maintains relatively stable fairness until absenteeism exceeds 50\%, indicating greater resilience across a broader range of scenarios.

Note that when absenteeism approaches $1$, delegates receive too little feedback from their delegators---who rarely vote---for effective preference learning. This shortage of training data contributes to the observed decline in fairness.

\paragraph{Cluster Density}
\begin{figure}[ht]
\centering
\includegraphics[width=1\textwidth]{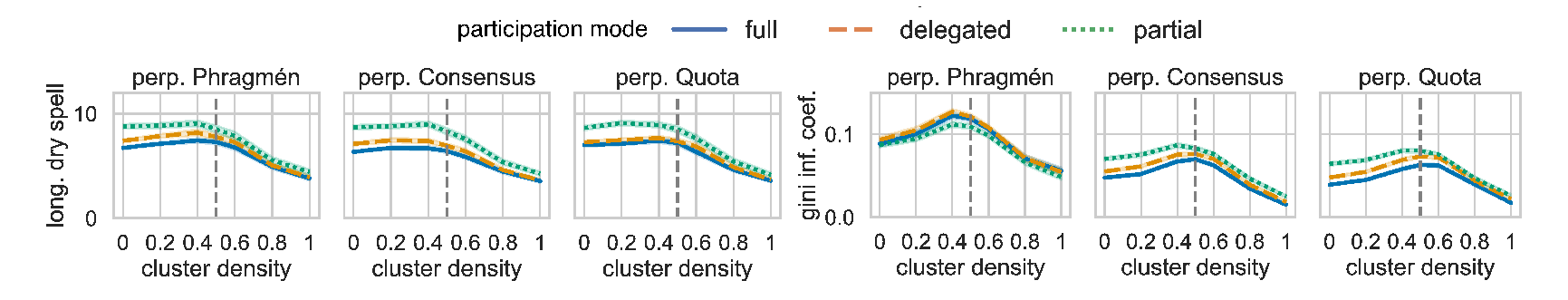}
\caption{Effects of cluster density on Gini influence coefficients and longest dry spells for three perpetual voting rules. Vertical dashed lines mark the default value, as used for \autoref{fig:boxplots_composition}. All other parameters are set to their defaults as defined in \autoref{sec:experiments}. 
}
\Description[Six line plots displaying the changes in fairness metrics as cluster densities vary, comparing different voting rules and participation modes.]{Line plots illustrate how fairness metrics—Longest Dry Spell and Gini Influence Coefficient—change with increasing cluster density. Both metrics remain relatively stable up to a certain density, after which they sharply decline for full, delegated, and partial turnout modes. Across most voting rules, partial turnout shows the highest metric values, delegated turnout is intermediate, and full turnout has the lowest values. The exception is Perpetual Phragmén, where partial turnout exhibits the lowest Gini Influence Coefficient among the three participation modes.}
\label{fig:variants_density}
\end{figure}
\autoref{fig:variants_density} shows the effect of cluster density.  Similar to polarization, higher density leads to greater homogeneity within both majority and minority groups, reducing variation in their approval sets.  This tends to improve fairness, as reflected by lower Gini influence coefficients and Longest Dry Spell values at high densities. As polarization increases, voter preferences converge around two distinct approval sets, allowing fairness to be maximized by alternating appropriately between them.

At the other extreme, low cluster density dissolves the clear majority–minority structure, slightly improving fairness by reducing majority dominance in the selection process. However, in the intermediate range, neither the benefits of diversity nor the clarity of polarization are present. The persistence of majority group influence, combined with still diverse approval sets, makes it harder to satisfy large portions of the electorate while maintaining fairness---resulting in increased Longest Dry Spell and Gini influence coefficient.

\paragraph{Minority Fraction}

\begin{figure}[ht]
\centering
\includegraphics[width=1\textwidth]{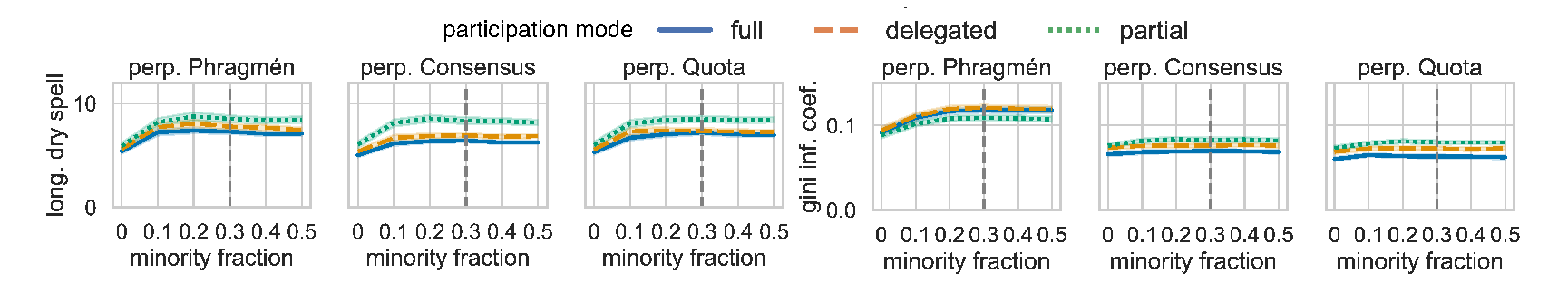}
\caption{Effects of the minority group's size on Gini influence coefficients and longest dry spells for three perpetual voting rules. Vertical dashed lines mark the default value, as used for \autoref{fig:boxplots_composition}. All other parameters are set to their defaults as defined in \autoref{sec:experiments}. 
}
\Description[Six line plots depicting the change in fairness metrics as minority fraction varies, comparing different voting rules and participation modes.]{Line plots show how the fairness metrics—Longest Dry Spell and Gini Influence Coefficient—change with increasing minority fraction. Both metrics steadily increase until the minority fraction reaches 0.1, after which they remain stable across full, delegated, and partial turnout modes. Generally, partial turnout yields the highest fairness values, followed by delegated, with full turnout having the lowest. An exception is seen in Perpetual Phragmén, where partial turnout results in the lowest Gini Influence Coefficient among the three modes.}
\label{fig:variants_minority}
\end{figure}
Beyond varying cluster density, the balance between majority and minority groups can also be adjusted by changing the size of the minority group, as shown in \autoref{fig:variants_minority}. When the minority fraction is 0, the electorate forms a single, highly cohesive group, resulting in improved fairness due to broad agreement among voters.

As the minority fraction increases from zero, fairness generally declines---particularly in terms of Longest Dry Spell. Smaller minority groups contribute less to overall representation demands, making it easier for the majority to dominate outcomes. While perpetual voting rules aim to balance representation over time, small minorities exert limited influence, leading to longer intervals before their preferences are satisfied.

This effect is less pronounced in terms of the Gini influence coefficient, since smaller minorities have proportionally smaller quotas. Their limited demands can therefore still be met without significantly disrupting overall balance, keeping Gini influence coefficients relatively stable even as Longest Dry Spells increase.

\paragraph{Noise}
\begin{figure}[ht]
\centering
\includegraphics[width=1\textwidth]{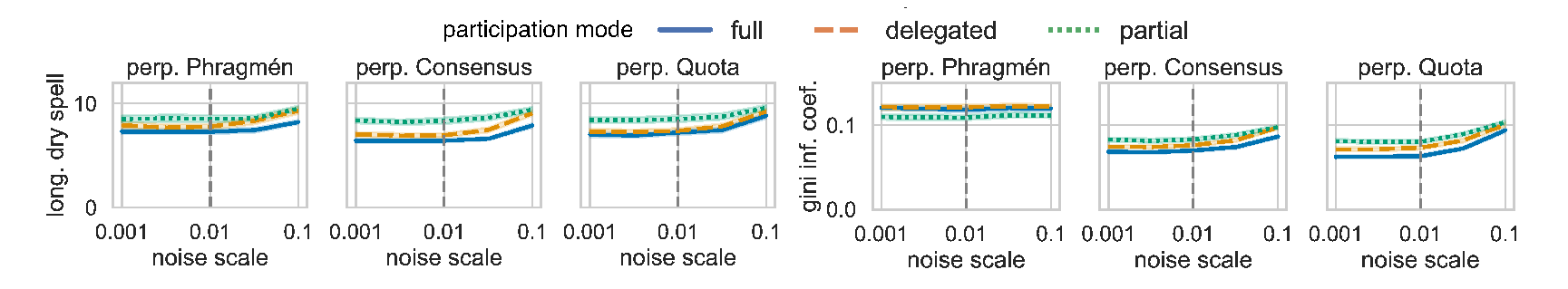}
\caption{Effects of voter noise on Gini influence coefficients and longest dry spells for three perpetual voting rules. Vertical dashed lines mark the default value, as used for \autoref{fig:boxplots_composition}. All other parameters are set to their defaults as defined in \autoref{sec:experiments}. 
}
\Description[Six line plots depicting the change in fairness metrics as noise varies, comparing different voting rules and participation modes.]{Line plots show that as noise increases, values of both fairness metrics, Longest dry spell and Gini influence coefficient, remain relatively stable, with a slight upward trend toward the highest noise levels. Partial turnout typically shows the highest values, followed by delegated, with full turnout lowest. Perpetual Phragmén differs slightly, where all participation modes remain stable across noise levels, and partial turnout has the lowest Gini Influence Coefficient among the three modes.}
\label{fig:variants_noise}
\end{figure}
In our random utility model, voters approve options when their utility---perturbed by zero-centered noise---exceeds a threshold. As shown in \autoref{fig:variants_noise}, increasing noise levels reduces fairness. This is because higher noise introduces greater variability in individual approvals, weakening group-level consistency and making preferences more personalized. The resulting dispersion in approval sets mirrors the effects observed with lower cluster density in \autoref{fig:variants_density}, driven by the same underlying mechanism: reduced alignment within voter groups makes it harder to achieve balanced and representative outcomes.

\subsection{Learning delegates preserve election outcomes}

\begin{figure}[h]
\centering
\includegraphics[width=.9\textwidth]{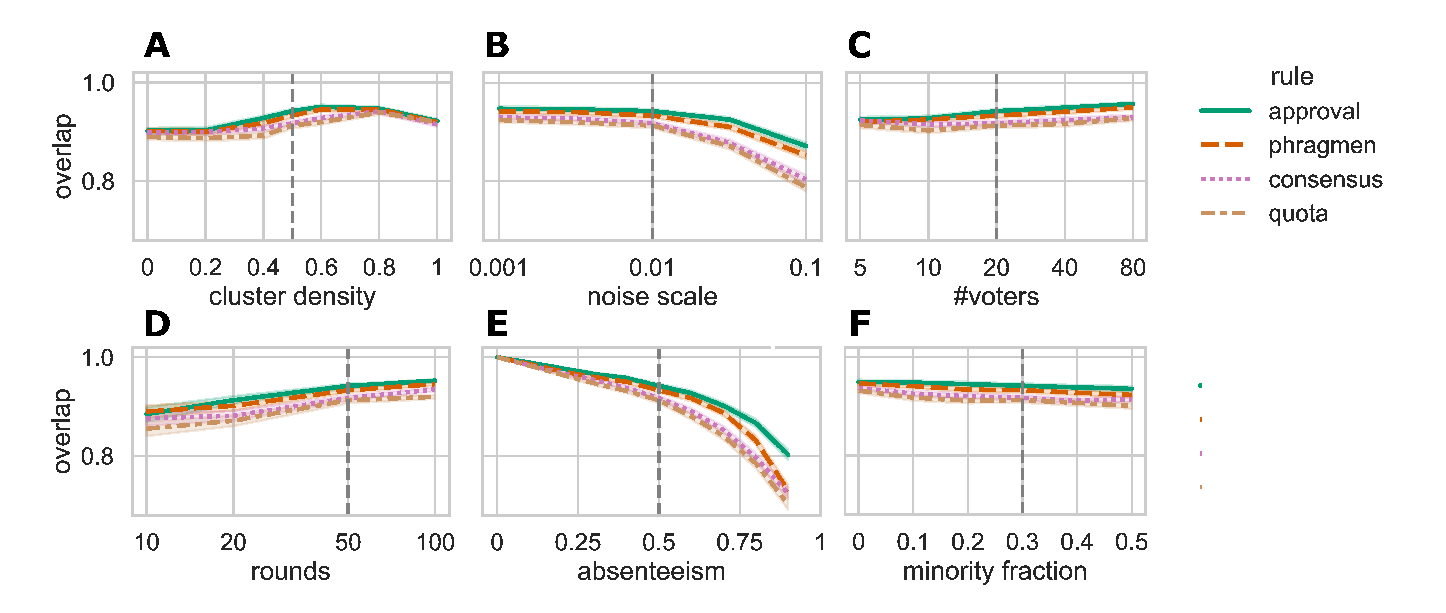}
\caption{Effects of parameter variations on overlap. We assess the impact across three voting rules for: \textbf{A.} cluster density, \textbf{B.} how noisy voters are, \textbf{C.} the number of rounds, \textbf{D.} the number of voters, \textbf{E.} how probable absenteeism is, and \textbf{F.} fraction of voters in the minority group. Vertical dashed lines mark the default value, as used for \autoref{fig:boxplots_composition}. All other parameters are set to their defaults as defined in \autoref{sec:experiments}. 
}
\Description[Six line plots illustrate how the ability of delegates to preserve election outcomes changes under varying parameters across voting rules.]{Line plots A, C, and D illustrate a slight increase in overlap with higher cluster density, a larger number of voters, and more voting rounds, respectively, while a sharp decline in overlap occurs as noise scale (plot B) and absenteeism (plot E) increase. When the minority fraction (plot F) varies, overlap remains comparatively constant across all voting rules.}
\label{fig:overlap_variants}
\end{figure}
In addition to fairness metrics, a key requirement for Artificial Delegates is that they do not distort election outcomes. Ideally, their participation should yield results close to what would have occurred under either full or partial turnout. We evaluate this aspect in terms of election overlap (See Section \ref{sec:overlap}), showing in \autoref{fig:overlap_variants} how much the votes with Artificial Delegates overlap with those that would have been obtained through full or partial turnout. 

The number of voters and the minority fraction have little effect on outcome overlap. In contrast, other parameters show more pronounced or nuanced influence. Increased absenteeism reduces available training data, which weakens delegate performance and lowers overlap. In contrast, increasing the number of rounds improves overlap by giving delegates more training data, leading to better models and decisions that more closely match full or partial turnout.
 Conversely, increasing the noise makes the training data less reliable, resulting in slightly lower overlap.

Finally, increasing cluster density makes voters’ underlying utilities more similar within groups. Although approval sets remain noisy, this alignment increases the likelihood that certain options are approved by a larger plurality within each group. These consistent signals make it easier for delegates to identify and reproduce group-level approval trends, thereby improving outcome overlap.

\section{Conclusion}

Absenteeism is an unavoidable aspect of real-world decision-making settings, such as participatory budgeting in small communities or corporate governance, yet most perpetual voting rules rely on consistent participation. When participants are absent from certain voting rounds, the balancing mechanisms of these rules are disrupted, leading to less proportional outcomes. To address this, we introduce Artificial Delegates---models that learn the preferences of absent voters based on their past voting behavior.

Our results show that, while all perpetual voting rules are vulnerable to absenteeism, the use of Artificial Delegates restores representation across key dimensions: influence is more equally distributed (lower Gini influence coefficients), minority voices are satisfied more regularly (shorter dry spells), and group entitlements are better respected (better quota compliance). Crucially, they also preserve outcome fidelity, producing results that align with those under full or partial turnout, thus ensuring artificial delegates rarely lead to outcomes which would not have won in their absence. We further show that this improvement in fairness and equal representation remains robust across a wide range of parameter settings. 

While this work serves as an initial exploration of the impacts of absenteeism, and the benefits of Artificial Delegates, future work should further consider practical aspects. 
We also assumed that individuals have stable preferences, whereas in reality, human preferences may evolve over time in response to new information, experiences, social feedback, and shifting priorities. For instance, in participatory budgeting,community members are likely to engage in discussions, negotiations, and be influenced by others' choices. To infer preferences, we assumed no communication or knowledge about others’ actions, meaning approval sets were strictly personalized. 

Moreover, when faced with a large number of alternatives or features, voters may experience cognitive overload, which can significantly impact their ability to process information and make informed votes. While our exploration of voter clusters and the random utility model can, to some degree, capture these concerns, the model assumes a linear utility function --- a simplification of the noisy and inconsistent nature of human decision making. Furthermore, the number of voter experiences (or historical data) available might also influence the effectiveness of an AI delegate. Addressing these challenges remains an important avenue for future research. Finally, future work could exploit the learned utilities in themselves, rather than only the approval sets we derive from them, to further enhance fairness in dynamic voting environments.

\bibliographystyle{ACM-Reference-Format}
\bibliography{a_references}


\end{document}